\documentclass{article}
\usepackage{spconf,amsmath,graphicx,hyperref}
\usepackage{times}
\usepackage{algorithm}
\usepackage{algorithmic}
\usepackage{booktabs}
\usepackage{multirow}
\title{NCV: A Node-Wise Consistency Verification Approach for Low-Cost Structured Error Localization in LLM Reasoning}
\name{
 \parbox{\linewidth}{\centering
Yulong Zhang$^{1,2,3}$, Li Wang$^{3}$, Wei Du$^{3,2}$, Peilin Li$^{4}$, Yuqin Dai$^{4}$ Zhiyuan Zhao$^{3}$, \\ 
Lingyong Fang$^{2,3}$, Ziniu Liu$^{3}$, Ru Zhang$^{1}$ Huijia Zhu$^{3}$, Gongshen Liu$^{2,5*}$\thanks{*: Corresponding author}}
}
\address{$^1$Beijing University of Posts and Telecommunications \quad $^2$Shanghai Jiao Tong University\\
$^3$Ant Group \quad $^4$Tsinghua University \quad $^5$Inner Mongolia Research Institute of SJTU}

\begin{document}
\ninept
\maketitle
\begin{abstract}
Verifying multi-step reasoning in large language models is difficult due to imprecise error localization and high token costs. Existing methods either assess entire reasoning chains, suffering attention dilution, or rely on expensive multi-sampling. We introduce Node-wise Consistency Verification (NCV), a training-free framework that recasts verification as lightweight binary consistency checks at the node level. By decomposing the chain of thought into interconnected verification nodes, NCV precisely localizes errors and avoids unnecessary long-form generation. Experiments demonstrate that our approach enhances interpretability and efficiency, presenting a scalable solution for reliable LLM reasoning verification. On public datasets, NCV achieves a 10\% to 25\% improvement in F1 scores over baselines while utilizing $6\times$~$58\times$ fewer tokens than traditional methods like CoT-based verifiers.

\end{abstract}

\section{Introduction}

Rapid advancement of large language models (LLMs) has led to unprecedented capabilities in complex problem solving~\cite{Achiam2023Gpt, Shao2024Deepseekmath, qwen2.5-math,chowdhery2022palm,verma2024ghostbuster,yin2025floorplan, jiang2025caporeinforcingconsistentreasoning}. However, ensuring the reliability of their multi-step reasoning remains a fundamental challenge in AI safety and trustworthiness. While Chain-of-Thought (CoT) prompting~\cite{wei2022chain} and related approaches like zero-shot reasoning~\cite{kojima2022large} and scratchpad methods~\cite{nye2021show} have significantly enhanced LLMs' reasoning capabilities, they introduce critical limitations that hinder practical deployment: excessive token consumption, extended context lengths that slow inference speed, and increased deployment costs. More concerning, LLMs frequently produce reasoning chains containing subtle errors that are difficult to identify and localize, particularly when the reasoning appears plausible but contains fundamental logical flaws~\cite{He2025CanLL, Luo2024Improve}.

The core challenges in reasoning verification stem from both interpretability and efficiency concerns. \textbf{Interpretability and Error Localization}: Current verification methods struggle to provide fine-grained error localization and lack interpretability in their decision-making process. When errors occur in multi-step reasoning, it becomes difficult to pinpoint exactly where the reasoning breaks down, limiting both debugging capabilities and trust in the system. \textbf{Attention Dilution}: End-to-End (E2E) verification methods attempt to validate entire reasoning chains simultaneously, but suffer from attention dilution when processing long sequences~\cite{Hong2023ACL, Stechly2024OnTS, Liu2023LostInTheMiddle, dai2025evinote}. As reasoning chains grow longer and more complex, the model's attention becomes scattered across numerous steps, reducing its ability to focus on critical logical dependencies and error-prone transitions. \textbf{Computational Inefficiency}: While Chain-of-Thought reasoning improves accuracy, it comes at a significant computational cost. The lengthy reasoning processes consume substantial tokens, leading to slower inference speeds and higher deployment costs, particularly problematic for large-scale applications requiring real-time responses.

Existing verification approaches fail to adequately address these challenges. Process Reward Models (PRMs) offer step-level assessment but require extensive supervised training and struggle with generalization to new problem types~\cite{Lightman2023LetsVS, Wang2023MathShepherdVA}. Recent decomposition strategies often rely on expensive multi-sampling techniques or complex premise extraction mechanisms that further increase computational overhead~\cite{mukherjee2025premiseaugmented}. Moreover, current language models are prone to making process errors even when reaching correct final answers, particularly on challenging mathematical problems, underscoring the limitations of outcome-based evaluation.

\begin{figure*}[t]

\centering
\includegraphics[width=1\textwidth]{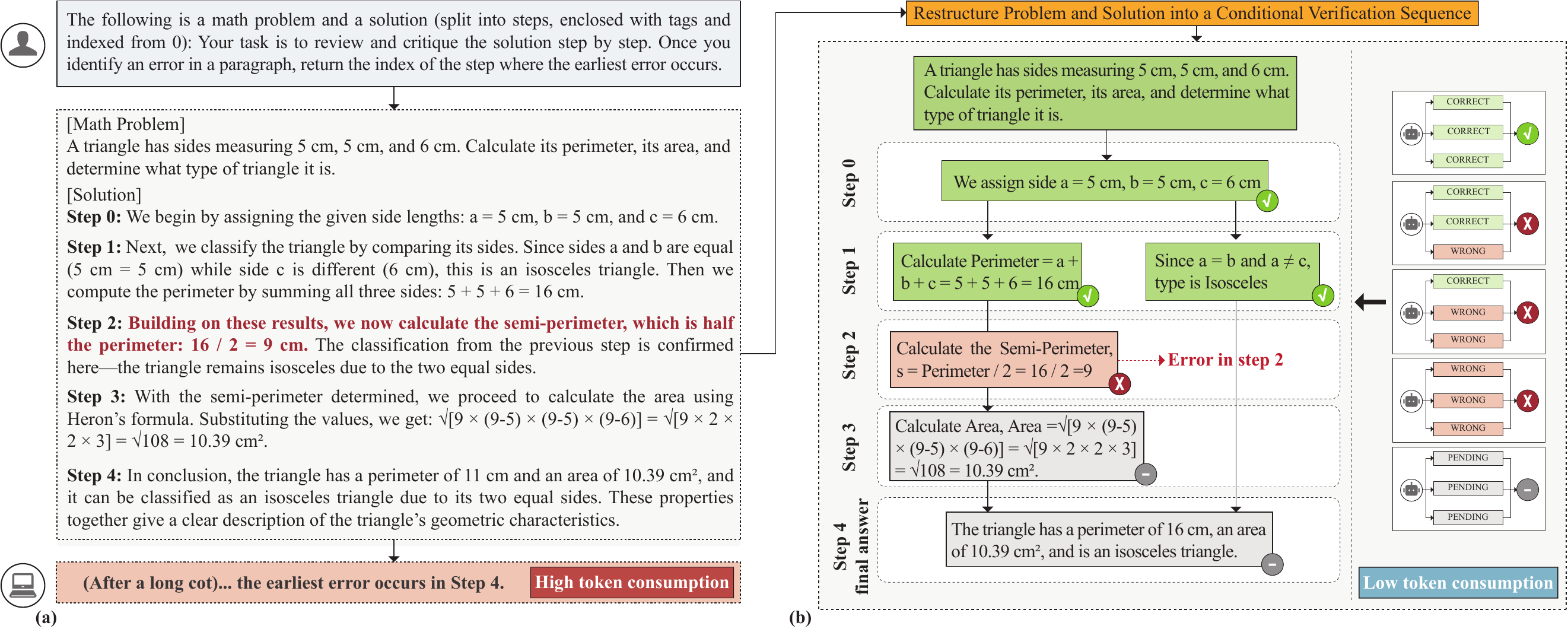}
\vspace{-1\baselineskip}
\caption{Comparison between End-to-End (E2E) verification and Node-wise Consistency Verification (NCV). (a) E2E approach processes the entire problem and solution simultaneously, consuming extensive tokens through Chain-of-Thought reasoning but failing to precisely localize errors. (b) NCV decomposes the reasoning into structured steps and nodes, enabling sequential verification with minimal token consumption and accurate error localization.}
\label{fig:main}
\end{figure*}
To address these challenges, we introduce \textbf{Node-wise Consistency Verification (NCV)}, a training-free framework that transforms complex reasoning verification through structured decomposition, as illustrated in Figure~\ref{fig:main}. Our key discovery is that converting Chain-of-Thought reasoning into structured decomposition dramatically improves verification effectiveness for long reasoning chains. Inspired by Self-Consistency~\cite{wang2022self} and Consensus Entropy~\cite{zhang2025consensusentropyharnessingmultivlm}, we further hypothesize that restructuring complex reasoning into multiple simple verification problems enables extremely low-cost consistency checks with superior error localization capabilities.

NCV restructures long reasoning chains into granular verification nodes, transforming a single complex verification task into multiple simple binary judgment problems. This approach enables precise error localization while consuming minimal computational resources through efficient consistency mechanisms.

We conduct comprehensive experiments on all four subsets of the ProcessBench benchmark (GSM8K~\cite{gsm8k}, MATH~\cite{math}, OlympiadBench~\cite{olympiadbench}, and Omni-MATH~\cite{omnimath}), comparing against both eight-sampling majority voting and greedy decoding baselines. Our results demonstrate that NCV achieves substantial performance improvements across all datasets, with F1 score gains ranging from 10-25\% while consuming 6-58× fewer tokens than conventional approaches. The primary contributions of this work include:

\begin{itemize}
    \item \textbf{Structured Decomposition Discovery}: We demonstrate that transforming Chain-of-Thought reasoning into structured decomposition significantly improves verification effectiveness for long reasoning chains, providing a new perspective on reasoning verification challenges.
    
    \item \textbf{Node-wise Consistency Verification Framework}: We propose NCV, which restructures complex long reasoning chains into multiple simple verification problems, enabling extremely low-cost consistency verification with exceptional effectiveness for error localization.
    
    \item \textbf{Comprehensive Performance and Cost Analysis}: We conduct extensive experiments comparing performance metrics and computational costs, demonstrating that NCV achieves superior precision verification effects at significantly lower costs than existing methods.
\end{itemize}

\section{Node-wise Consistency Verification}

Figure~\ref{fig:main} illustrates the core difference between conventional end-to-end verification and our proposed NCV approach. While end-to-end methods process entire reasoning chains holistically, NCV enables precise error localization through structured decomposition.

\subsection{Problem Formulation}

Given a problem $P$ and a solution $S = \{s_1, s_2, \ldots, s_n\}$, the verification task seeks a function $V: (P, S) \rightarrow \{0, 1, \ldots, n\}$ where:
\begin{equation}
V(P, S) = \begin{cases}
0 & \text{if } S \text{ correctly solves } P \\
i & \text{if step } s_i \text{ is the first incorrect step}
\end{cases}
\end{equation}

\textbf{End-to-End Baseline:} Conventional methods compute:
\begin{equation}
V_{\text{E2E-cot}}(P, S) = \text{LLM}_{\text{CoT}}(P \oplus S)
\end{equation}
where $\oplus$ denotes concatenation and $\text{LLM}_{\text{CoT}}$ generates lengthy reasoning consuming $O(|P| + |S|)$ tokens.

\subsection{Structured Decomposition}

The key insight of NCV is to transform the sequential reasoning chain into a structured conditional framework. Rather than treating the solution as a monolithic sequence, we decompose it into atomic verification units that can be independently validated.

NCV restructures $S$ into a conditional verification sequence $\mathcal{N} = \{n_1, \ldots, n_m\}$. Common structures include: (1) linear chain structure for sequential reasoning, (2) single-source, single-sink directed acyclic graph for complex dependencies, and (3) other hybrid structures as needed. All preserve logical flow while enabling granular verification.

\textbf{Step-to-Node Mapping:} Each reasoning step $s_i$ decomposes into atomic assertions:
\begin{equation}
s_i \rightarrow \{n_{i,1}, n_{i,2}, \ldots, n_{i,k_i}\}
\end{equation}
In many cases, a step contains only a single atomic assertion ($k_i = 1$), significantly simplifying the verification task. This fine-grained decomposition transforms complex reasoning verification into simple factual checks.

\textbf{Flexible Conditional Structure:} The specific structure depends on the reasoning pattern. When clear logical dependencies exist, we can construct explicit edges; otherwise, we default to a linear conditional chain where each node $n_i$ conditions on the problem $P$ and all previously verified nodes. This flexibility allows NCV to adapt to diverse reasoning styles while maintaining verification effectiveness.

\subsection{Sequential Node Verification}

The core advantage of our approach lies in transforming complex verification into simple conditional judgments. For each node in the structured sequence, we define the verification probability:
\begin{equation}
P(\text{correct}(n_i) | \text{PriorSteps}_i)
\end{equation}
where $\text{PriorSteps}_i = P \cup \{n_j : j < i, \text{verified}(n_j) = \text{true}\}$ contains the problem and all previously verified nodes.

\textbf{Binary Mode:} By default, NCV restricts model output to binary judgments, dramatically reducing complexity and enabling efficient verification:
\begin{equation}
V_{\text{NCV}}(n_i) = \text{LLM}_{\text{binary}}(\text{correct}(n_i) | \text{PriorSteps}_i)
\end{equation}
The fine-grained step-to-node decomposition ensures that most verification tasks become simple factual checks, eliminating the need for lengthy reasoning chains. This binary approach offers multiple advantages: significantly reduced token consumption, faster inference speed, and compatibility with smaller, non-reasoning models, resulting in substantially lower computational costs.

\textbf{Reasoning Mode:} When computational budget permits, we can allow full reasoning chains:
\begin{equation}
V_{\text{reasoning}}(n_i) = \text{LLM}_{\text{CoT}}(\text{verify}(n_i) | \text{PriorSteps}_i)
\end{equation}
This mode requires more capable models and higher token consumption but reduces the need for consistency mechanisms, as the model can provide detailed justifications for its judgments.

\subsection{Consistency Strategies for Binary Mode}

When using binary mode, we employ consistency mechanisms to enhance reliability. The intuition is that simple verification tasks benefit from multiple independent judgments rather than elaborate reasoning. Since atomic nodes represent simple factual checks, multiple binary judgments are both computationally efficient and highly effective.

\textbf{Multi-Sampling Voting:} We generate $k$ independent binary judgments and use majority voting:
\begin{equation}
V_{\text{vote}}(n_i) = \text{Majority}\left(\{V_{\text{NCV}}^{(j)}(n_i)\}_{j=1}^k\right)
\end{equation}

\textbf{One-Vote Veto:} For conservative error detection, any single incorrect judgment flags the node as erroneous:
\begin{equation}
V_{\text{veto}}(n_i) = \begin{cases}
\text{Incorrect} & \text{if } \exists j: V_{\text{NCV}}^{(j)}(n_i) = \text{Incorrect} \\
\text{Correct} & \text{otherwise}
\end{cases}
\end{equation}

Note that consistency strategies are primarily needed in binary mode. In reasoning mode, the model's detailed justifications typically provide sufficient confidence for single-shot verification.

\subsection{NCV Algorithm}

The NCV verification process follows a simple sequential procedure:

\begin{algorithm}[h]
\caption{Node-wise Consistency Verification}
\begin{algorithmic}[1]
\REQUIRE Problem $P$, Solution $S$, Node sequence $\mathcal{N} = \{n_1, \ldots, n_m\}$
\ENSURE Verification result: 0 (correct) or error location $i$
\STATE Order $\leftarrow$ StructuralSort($\mathcal{N}$)
\FOR{each $n_i$ in Order}
    \STATE PriorSteps$_i \leftarrow P \cup \{n_j : j < i, \text{verified}(n_j)\}$
    \STATE Result$_i \leftarrow$ ConsistencyVerify($n_i$, PriorSteps$_i$)
    \IF{Result$_i$ = Incorrect}
        \RETURN StepIndex($n_i$)
    \ENDIF
\ENDFOR
\RETURN 0
\end{algorithmic}
\end{algorithm}

\subsection{Computational Efficiency Analysis}
NCV improves efficiency by decomposing complex verification into simple binary checks. Token costs: End-to-end (CoT): \(\text{Cost}_{\text{E2E-cot}} = O(|P|+|S|)\,C_{\text{reasoning}}\); NCV-binary: \(\text{Cost}_{\text{binary}} = m\,k\,C_{\text{binary}}\), where \(m\) is the number of nodes (claims) and \(k\) the average checks per node. Since \(C_{\text{binary}} \ll C_{\text{reasoning}}\), typically \(\text{Cost}_{\text{binary}} \ll \text{Cost}_{\text{E2E-cot}}\) for moderate \(m k\), enabling substantial savings and the use of smaller, cost-effective models for large-scale deployment.






\section{Experiments}

We evaluate NCV on ProcessBench~\cite{Zheng2024ProcessBenchIP}, focusing on accuracy, error localization, and computational efficiency across different model and model sizes and reasoning complexity levels.

\begin{table*}[!htbp]
  \centering
  \caption{Comparison on ProcessBench. We report Correct Accuracy (accuracy of successfully identifying completely correct solutions), Error Locating accuracy (accuracy of successfully locating error positions), and F1 scores.}
  \label{tab:performance_comparison}
  \resizebox{\linewidth}{!}{
  \begin{tabular}{llccccccccccccc}
  \specialrule{1.5pt}{0pt}{0pt}
  \multirow{2}{*}{\textbf{Model}} & \multirow{2}{*}{\textbf{Method}} & \multicolumn{3}{c}{\textbf{GSM8K}} & \multicolumn{3}{c}{\textbf{MATH}} & \multicolumn{3}{c}{\textbf{OlympiadBench}} & \multicolumn{3}{c}{\textbf{Omni-MATH}} & \multirow{2}{*}{\textbf{Avg F1}} \\
  \cmidrule(lr){3-5} \cmidrule(lr){6-8} \cmidrule(lr){9-11} \cmidrule(lr){12-14}
   &  & Correct & Error & \textbf{F1} & Correct & Error & \textbf{F1} & Correct & Error & \textbf{F1} & Correct & Error & \textbf{F1} & \\ \specialrule{1.15pt}{0pt}{0pt}
  \multirow{3}{*}{Qwen2.5-7B} & E2E-cot (8-vote) & 33.2 & 40.6 & 36.5 & 45.1 & 30.8 & 36.6 & 33.9 & 26.5 & 29.7 & 28.6 & 26.2 & 27.4 & 32.6 \\
  & E2E-cot (greedy) & 66.3 & 36.7 & 47.3 & 63.8 & 23.7 & 34.6 & 46.0 & 25.4 & 32.7 & 43.6 & 26.1 & 32.6 & 36.8 \\
  & NCV@3-B (ours) & \textbf{85.5} & \textbf{39.1} & \textbf{53.7} & \textbf{68.0} & \textbf{38.9} & \textbf{49.5} & \textbf{52.2} & \textbf{25.9} & \textbf{34.6} & \textbf{57.3} & \textbf{27.0} & \textbf{36.7} & \textbf{43.6} \\
  \hline
  \multirow{3}{*}{Qwen2.5-32B} & E2E-cot (8-vote) & 97.9 & 49.3 & 65.6 & 95.8 & 36.7 & 53.1 & 95.9 & 25.3 & 40.0 & 92.5 & 24.1 & 38.3 & 49.3 \\
  & E2E-cot (greedy) & 97.9 & 43.0 & 59.8 & 95.6 & 33.3 & 49.4 & 90.0 & 22.4 & 35.9 & 87.6 & 22.4 & 35.7 & 45.2 \\
  & NCV@3-B (ours) & 94.8 & \textbf{67.6} & \textbf{78.9} & 83.3 & \textbf{66.7} & \textbf{74.0} & 69.3 & \textbf{55.8} & \textbf{61.9} & 67.6 & \textbf{55.9} & \textbf{61.2} & \textbf{69.0} \\
  \hline
  \multirow{3}{*}{Qwen2.5-72B} & E2E-cot (8-vote) & 96.9 & 62.8 & 76.2 & 93.1 & 46.3 & 61.8 & 92.6 & 38.7 & 54.6 & 90.9 & 36.6 & 52.2 & 61.2 \\
  & E2E-cot (greedy) & 98.4 & 61.4 & 75.6 & 91.9 & 45.3 & 60.7 & 88.5 & 33.7 & 48.9 & 88.4 & 33.7 & 48.8 & 58.5 \\
  & NCV@3-B (ours) & 96.4 & \textbf{62.3} & 75.7 & 83.0 & \textbf{55.1} & \textbf{66.2} & 74.3 & \textbf{44.8} & \textbf{55.9} & 73.0 & \textbf{44.7} & \textbf{55.4} & \textbf{63.3} \\
  \hline
  \multirow{3}{*}{Llama-3.3-70B} & E2E-cot (8-vote) & 96.9 & 72.5 & \textbf{82.9} & 94.6 & 43.3 & 59.4 & 94.1 & 31.0 & 46.7 & 90.5 & 28.2 & 43.0 & \textbf{58.0} \\
  & E2E-cot (greedy) & 96.9 & 66.2 & 78.6 & 93.1 & 38.4 & 54.4 & 90.0 & 30.9 & 46.0 & 86.3 & 27.1 & 41.3 & 55.1 \\
  & NCV@3-B (ours) & 92.2 & 57.0 & 70.5 & 78.3 & \textbf{47.6} & \textbf{59.5} & 54.3 & \textbf{49.0} & \textbf{51.6} & 64.7 & \textbf{41.6} & \textbf{50.7} & \textbf{58.0} \\
  \specialrule{1.5pt}{0pt}{0pt}
  \end{tabular}
  }
\end{table*}

\subsection{Experimental Setup}

\textbf{Dataset:} We evaluate our proposed NCV framework on ProcessBench~\cite{Zheng2024ProcessBenchIP}, a comprehensive benchmark designed to assess the ability to identify erroneous steps in mathematical reasoning. ProcessBench consists of 3,400 test cases across four subsets with varying difficulty levels: GSM8K (400 cases, elementary-level word problems), MATH (1,000 cases, competition-level mathematics), OlympiadBench (1,000 cases, Olympiad-level problems), and Omni-MATH (1,000 cases, advanced mathematical reasoning across diverse domains). Each test case contains a step-by-step solution with error location annotated by multiple human experts to ensure reliability. The task requires models to identify the first wrong step or conclude that all steps are correct if no errors exist.

\textbf{Baselines:} We compare against E2E-cot methods: (1) \textit{E2E-cot (8-vote)} using majority voting across eight chains, and (2) \textit{E2E-cot (greedy)} with single-pass verification.

\textbf{Implementation:} NCV@3-Binary uses three-fold consistency checking with binary judgments, constraining output to 4 tokens maximum for efficiency.

\subsection{Main Results}

Table~\ref{tab:performance_comparison} presents comprehensive results comparing our NCV framework against strong baselines across four ProcessBench subsets. The results demonstrate several key findings that validate our approach.

\textbf{Consistent Superior Performance:} NCV achieves superior F1 scores across all model sizes and datasets, with particularly notable improvements in error localization accuracy. The consistency of these gains across different model architectures (Qwen2.5 and Llama-3.3) demonstrates the generalizability of our approach. Even for the largest models where baseline performance is already high, NCV continues to provide meaningful improvements.

\textbf{Scaling Benefits with Problem Complexity:} The performance gains of NCV increase with problem difficulty. For Qwen2.5-32B, we observe F1 improvements of 13.3 points on GSM8K (elementary-level), 20.9 points on MATH (competition-level), 21.9 points on OlympiadBench (Olympiad-level), and 23.8 points on Omni-MATH (advanced reasoning). This trend indicates that NCV's structured decomposition approach becomes increasingly valuable for complex reasoning tasks where E2E methods struggle with attention dilution.

\textbf{Superior Error Localization:} NCV demonstrates particularly strong performance in error localization, which is crucial for practical applications. While E2E methods often correctly identify that an error exists but fail to pinpoint its location, NCV's node-wise verification enables precise error identification. For example, with Qwen2.5-32B on OlympiadBench, NCV achieves 55.8\% error localization accuracy compared to 25.3\% for E2E-cot (8-vote).

\textbf{Efficiency vs. Accuracy Trade-off:} Comparing the two baselines reveals the classic trade-off between computational cost and accuracy. E2E-cot (8-vote) generally outperforms E2E-cot (greedy) but requires 8× more computation. Remarkably, NCV achieves better performance than both baselines while using significantly fewer tokens than the 8-vote approach, effectively breaking this trade-off.
\vspace{-0.2cm}
\subsection{Ablation Experiments}

We systematically analyze different verification strategies to understand the contribution of key components in our NCV framework. All experiments are conducted using Qwen2.5-32B-Instruct for consistent comparison.

\vspace{-0.3cm}

\begin{table}[!htbp]
  \centering
  \caption{Ablation study on ProcessBench using Qwen2.5-32B-Instruct. We report average F1 across all four subsets and relative performance vs. E2E-cot (greedy).}
  \label{tab:ablation}
  \begin{tabular}{lccc}
  \specialrule{1.5pt}{0pt}{0pt}
  \textbf{Method} & \textbf{Components} & \textbf{Avg F1} & \textbf{Rel.} \\
  \specialrule{1.15pt}{0pt}{0pt}
  E2E-CoT (greedy) & w/o Str\&Con & 45.2 & 100\% \\
  E2E-CoT (3-vote) & w/o Structure & 48.3 & 106\% \\
  NCV@1-Binary & w/o Consistency & 54.9 & 121\% \\
  \hline
  NCV@3-Binary & All Components & \textbf{61.4} & \textbf{136\%} \\
  NCV@3-cot & All Components & \textbf{69.0} & \textbf{153\%} \\
  \specialrule{1.5pt}{0pt}{0pt}
  \end{tabular}
\end{table}

\textbf{Each Component Helps:} Relative to E2E-CoT (greedy, 45.2), adding simple voting gives a modest gain (48.3; 106\%), adding consistency alone yields a larger boost (NCV@1-Binary: 54.9; 121\%), and combining structured decomposition with consistency further improves performance (NCV@3-Binary: 61.4; 136\%).

\textbf{NCV@3-CoT Uses CoT Per Node:} NCV@3-CoT applies Chain-of-Thought reasoning at every verification node and achieves the best F1 (69.0; 153\%). This demonstrates that NCV can trade additional tokens for higher accuracy when budget allows~\cite{wei2022chain, wang2022self}.

\textbf{Balanced Efficiency:} NCV@3-Binary offers a strong efficiency–accuracy trade-off, while NCV@3-CoT shows the upper bound achievable by increasing token consumption within the same node-wise verification framework.

\subsection{Cost-Effectiveness Analysis}

We analyze the cost-effectiveness of different verification strategies by examining the relationship between computational cost and performance. Table~\ref{tab:cost_analysis} presents detailed token consumption and inference characteristics for each method.

\vspace{-0.3cm}
\begin{table}[!htbp]
  \centering
  \caption{Cost-effectiveness analysis using Qwen2.5-32B-Instruct.}
  \label{tab:cost_analysis}
  \begin{tabular}{lccc}
  \specialrule{1.5pt}{0pt}{0pt}
  \textbf{Method} & \textbf{F1 Score} & \textbf{Tokens} & \textbf{Max Len} \\
  \specialrule{1.15pt}{0pt}{0pt}
  E2E-cot (greedy) & 45.2 & 177.4 & 756 \\
  E2E-cot (8-vote) & 49.3 & 1619.2 & 2008 \\
  \hline
  NCV@3-Binary (ours) & \textbf{61.4} & \textbf{28.1} & \textbf{4} \\
  \specialrule{1.5pt}{0pt}{0pt}
  \end{tabular}
\end{table}

\vspace{-0.2cm}

\textbf{Exceptional Cost-Effectiveness:} NCV@3-Binary demonstrates remarkable cost-effectiveness, delivering 61.4\% F1 performance with only 28.1 tokens per sample on average. This represents a 6.3× token reduction compared to E2E-cot (greedy) while simultaneously improving F1 by 16.2 points. Compared to E2E-cot (8-vote), the efficiency gain is even more dramatic: 57.6× fewer tokens while achieving 12.1 points higher F1 score.

\textbf{Inference Speed Advantages:} The constrained output format of NCV enables significantly faster inference. With a maximum output length of just 4 tokens compared to 756-2008 tokens for E2E methods, NCV reduces both generation time and memory requirements. This makes NCV particularly suitable for real-time applications and large-scale deployment scenarios.

\textbf{Breaking the Accuracy-Efficiency Trade-off:} Traditional approaches face a fundamental trade-off between accuracy and efficiency. E2E-cot (8-vote) improves accuracy over greedy decoding by only 4.1 F1 points but requires 8× more computation. In contrast, NCV achieves 16.2 points improvement over greedy decoding with minimal computational overhead, effectively breaking this trade-off.

\textbf{Scalability Implications:} The token efficiency of NCV has significant implications for large-scale deployment. For a system processing 1M verification requests daily, NCV would consume approximately 28M tokens compared to 177M tokens for greedy E2E and 1.6B tokens for 8-vote E2E, resulting in substantial cost savings while providing superior accuracy.

\section{Conclusion}

We present NCV, a simple training-free framework that verifies reasoning by node-wise checks. On ProcessBench, NCV yields consistent F1 gains over E2E while using far fewer tokens. Ablations confirm that decomposition and consistency both matter; combining them works best. NCV@3-Binary is a practical default, and NCV@3-CoT trades extra tokens for higher accuracy when budget allows.

\vfill\pagebreak




\bibliographystyle{IEEEbib}
\bibliography{refs}

\end{document}